\documentclass[sigconf]{acmart}

\usepackage{xspace}
\usepackage{algorithm}
\usepackage{algorithmic}
\usepackage{multirow}
\usepackage{graphicx}
\usepackage{subfigure}
\usepackage{bbold} 
\usepackage{mathtools}
\usepackage{enumitem}
\usepackage{balance}
\usepackage{hyperref}

\def\onedot{\ifx\lettoken.\else.\null\fi\xspace}
\def\ie{\emph{i.e}\onedot} 

\newcommand{\eg}{\emph{e.g}\onedot} 

\AtBeginDocument{%
  \providecommand\BibTeX{{%
    \normalfont B\kern-0.5em{\scshape i\kern-0.25em b}\kern-0.8em\TeX}}}

\setcopyright{acmcopyright}
\copyrightyear{2022}
\acmYear{2022}

\begin{document}


\title[Perceptual Conversational Head Generation with Regularized Driver and Enhanced Renderer]{Perceptual Conversational Head Generation \\ with Regularized Driver and Enhanced Renderer}

\author{Ailin Huang}
\authornote{Both authors contributed equally to this work.}
\affiliation{%
  \institution{Megvii Research \\ Wuhan University}
}
\email{huangailin@megvii.com}

\author{Zhewei Huang}
\authornotemark[1]
\affiliation{%
    \institution{Megvii Research \\
    ~\\}
}
\email{huangzhewei@megvii.com}

\author{Shuchang Zhou}
\affiliation{%
    \institution{Megvii Research \\
    ~\\}
}
\email{zsc@megvii.com}

\renewcommand{\shortauthors}{Ailin Huang, Zhewei Huang, \& Shuchang Zhou}

\begin{abstract}
This paper reports our solution for ACM Multimedia ViCo 2022 Conversational Head Generation Challenge, which aims to generate vivid face-to-face conversation videos based on audio and reference images. Our solution focuses on training a generalized audio-to-head driver using regularization and assembling a high-visual quality renderer. We carefully tweak the audio-to-behavior model and post-process the generated video using our foreground-background fusion module. We get first place in the listening head generation track and second place in the talking head generation track on the official leaderboard. Our code is available at \url{https://github.com/megvii-research/MM2022-ViCoPerceptualHeadGeneration}.

\end{abstract}


\copyrightyear{2022}
\acmYear{2022}
\setcopyright{acmcopyright}\acmConference[MM '22]{Proceedings of the 30th ACM International Conference on Multimedia}{October 10--14, 2022}{Lisboa, Portugal}
\acmBooktitle{Proceedings of the 30th ACM International Conference on Multimedia (MM '22), October 10--14, 2022, Lisboa, Portugal}
\acmPrice{15.00}
\acmDOI{10.1145/3503161.3551577}
\acmISBN{978-1-4503-9203-7/22/10}



\maketitle

\section{Introduction}
In face-to-face communication, people can observe real-time expressions and demeanor and more accurately capture their counterpart's feelings. It is interesting to understand this communication behavior and generate vivid talking head videos using computer vision technology. Proper responsive listening behavior is essential as well to effective communication, and also of critical importance to make digital humans more realistic during face-to-face human-computer interaction and animation production.

Generating conversational head videos is challenging because it involves not only the processing of serialized speech signals but also video synthesis~\cite{chen2020talking}. We need to use the voice signal to infer the changes in the expressions and lip shapes of the people in the conversation. Then we also need to synthesize high-quality generated video frames based on the reference images of speakers and listeners. The prior art~\cite{suwajanakorn2017synthesizing} has shown that realistic digital humans can be generated from a large number of videos of the same speaker. However, further research is needed to build a digital human system for any speaker with less available digital information.

In our paper, we focus on training an audio-to-head driver with limited data and assembling a powerful renderer to generate vivid videos. Our main techniques include: 

\begin{itemize}
    \item We apply several neural network training techniques to improve the performance of audio-to-head driver training on limited data, and further explore ensemble learning to make the model more robust. 
	\item We assemble an enhanced renderer for producing visually better and more stable generated videos.
\end{itemize}

\begin{figure*}[tb]
	\centering
	\includegraphics[width=16cm]{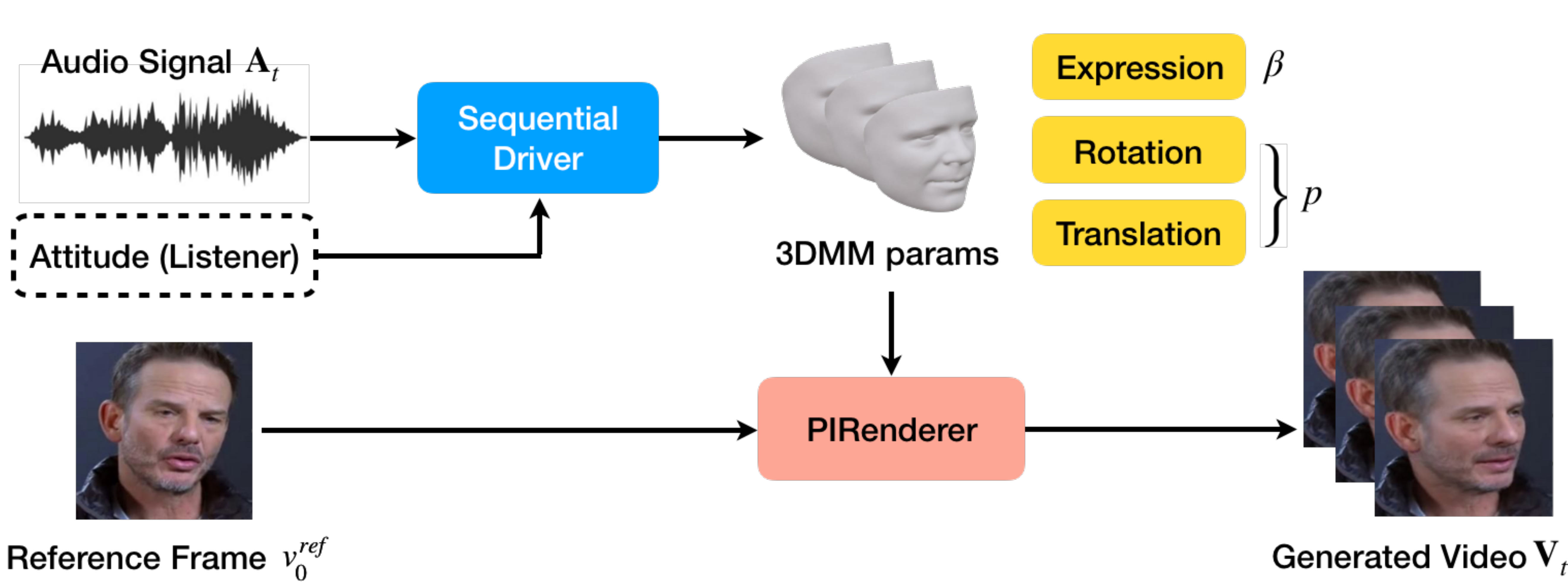}
	\caption{\textbf{Overview of video-driven head generation pipeline.} Given an audio signal sequence, a sequential driver approximates the 3DMM parameters for every video frame. Then a pre-trained PIRenderer~\cite{ren2021pirenderer} renders the final video based on these parameters and reference frame }\label{fig:main}
\end{figure*}

\section{Related Work}
\noindent\textbf{Audio-Driven Head Synthesis} 

With a reference head image and audio streams, many methods explore synthesizing a video based on a reference image, considering the head motion, expression, and lip shape changes. Some straightforward methods directly model the relationship between audio streams and reference images~\cite{jamaludin2019you,zhou2019talking}. To better correlate the video and audio factors, many recent works use a two-stage strategy to map reference images and audios firstly to feature representations~(\eg landmarks~\cite{chen2019hierarchical}, 3D morphable models~(3DMM)~\cite{thies2020neural,yi2020audio}, face parsing), and then render the final videos.

Recently, some very popular and advanced techniques have been applied to this task. StyleRig~\cite{tewari2020stylerig} describes a method to control StyleGAN~\cite{karras2019style} via a 3DMM. StyleGAN-based methods are very visually attractive, but their inference efficiency and identity retention performance still need improvement. The neural radiance field AD-NeRF~\cite{guo2021ad} can show very promising results for a scenario where the data of the target person is relatively sufficient.

\noindent\textbf{Neural Model and Regularization}

The head generation task requires a special collection of processed data. Training effective models with limited data is an important topic of the ViCo challenge~\cite{ViCo}. Recently, deep neural network with residual learning~\cite{he2016deep} has shown good performance in various tasks~\cite{lim2017enhanced,teed2020raft,huang2020rife,ding2021repvgg}. Meanwhile, many regularization techniques have been proposed to increase the generalization of neural networks~\cite{vincent2008extracting}. Among these techniques, Dropout~\cite{srivastava2014dropout} and Batch Normalization~(BN)~\cite{ioffe2015batch} are the most popular and powerful.

Besides, ensemble learning~\cite{dietterich2000ensemble} is a well-explored technique to improve the accuracy and generalization of the model. So it is usually used in various task, such as image classification~\cite{wortsman2022model}, reinforcement learning~\cite{kidzinski2018learning} and combination optimization~\cite{cao2022ml4co}. 

\section{Task Overview}

\subsection{Definition}
Our work uses a unified framework to learn both following tasks. We briefly introduce the definitions of these two tasks. The main results and experiments of this paper are presented on the talking head generation task.

\noindent\textbf{Vivid Talking Head Video Generation} 

Given the an input audio signal sequence $\mathbf{A}_t = {a_1, ..., a_t}$ of the speaker in time stamps ranging from $\{1, ..., t\}$ and a reference image $v_0^{ref}$, our goal is to generate a talking head video $\mathbf{V}^{Talking}_t = \{v_0, v_1, ..., v_t\}$.

\noindent\textbf{Responsive Listening Head Video Generation}

In addition to the input of the Talking Head Video Generation, the listening head generation additionally receives the input of the listener's attitude. Our goal is to generate a listening head video $\mathbf{V}^{Listening}_t = \{v_0, v_1, ..., v_t\}$.

\subsection{Dataset}
ViCo dataset~\cite{zhou2021responsive} contains $483$ video clips of $76$ listeners responding to $67$ speakers. The total length of these clips is approximately $95$ minutes. Following previous work~\cite{deng2019accurate, ren2021pirenderer}, we extract 3DMM parameters for each frame. As ViCo baseline~\cite{zhou2021responsive}, relative dynamic and identity-independent features face can be represented parametrically using $\{\beta \in \mathbb{R}^{64}, p \in \mathbb{R}^6\}$ which denotes the expression and pose. Here, $p$ represents rotations with $SO(3)\in \mathbb{R}^3$
and translations in $\mathbb{R}^3$. Additionally, for better modeling head movements, the baseline uses a ``crop" parameter $c$ of $\mathbb{R}^3$. This annotates where we will place and size the parametric 3D face in the original image.

\subsection{Evaluation}
We consider evaluating our models in terms of image quality and semantics. Because the generated images and the real video are unlikely to be pixel-by-pixel aligned, the traditional metric~(PSNR, SSIM) may not be reasonable. We mainly consider a metric at the feature level, Fréchet Inception Distance~(FID)~\cite{heusel2017gans}. We further analyze the landmark distance~(LMD) and expression feature distance~(ExpFD) between generated faces and ground truth. The whole system can render $4$ frames per second at $256\times 256$ resolution. Comparing our model with methods from other teams shown in Table~\ref{tab:leaderboard}, our model mainly gains an advantage on the FID index.

\begin{table}[t]
	\caption{\textbf{The official final leaderboard in ViCo challenge~\cite{ViCo}. } We intercept the results of the top few teams
	}
\label{tab:leaderboard}
\small
\resizebox{0.4\textwidth}{!}{\begin{tabular}{lccc}
\hline
\multicolumn{1}{c}{Team~(Talking)}                         & \multicolumn{1}{c}{PSNR$\uparrow$}   & \multicolumn{1}{c}{FID$\downarrow$}     & \multicolumn{1}{c}{LMD$\downarrow$}            \\ \hline 
\emph{sysu\_hcp} & \textbf{17.767} & 29.709 & \textbf{10.101} \\ 
\textbf{Ours} & 17.179 & \textbf{24.678} & 10.646 \\
\emph{iLearn} & 16.546 & 25.050 & 10.900 \\
\emph{Avatar} & 17.696 & 34.571 & 11.643 \\
\emph{Digital\_Human} & 17.331 & 30.944 & 12.837 \\
\emph{THU-Talking} & 16.390 & 45.361 & 12.707 \\
\hline
\multicolumn{1}{c}{Team~(Listening)}                         & \multicolumn{1}{c}{PSNR$\uparrow$}   & \multicolumn{1}{c}{FID$\downarrow$}     & \multicolumn{1}{c}{ExpFD$\downarrow$}            \\ \hline 
\textbf{Ours} & \textbf{18.512} & \textbf{21.350} & \textbf{0.116} \\
\emph{iLearn} & 18.491 & 26.675 & 0.133 \\
\emph{cheese} & 16.202 & 42.019 & 0.137 \\
\emph{en\_train} & 16.780 & 80.538 & 0.161 \\
\emph{LIMMC} & 16.265 & 86.983 & 0.167 \\
\hline
\normalsize
\vspace{-1.5em}
\end{tabular}}
\end{table}

\section{Method}
\subsection{Framework}
We illustrate the overall pipeline in Figure~\ref{fig:main}. We use a two-stage method to generate the head video with 3DMM parameters as the mediator interface. We firstly approximate the parameters of every frame using a sequential driver model. We then use a pre-trained PIRenderer~\cite{ren2021pirenderer} to render the final video based on these parameters and the reference image. Furthermore, we enhance PIRenderer use an image boundary inpainting trick and a foreground-background fusion module.

\begin{figure}[tb]
	\centering
	\includegraphics[width=5cm]{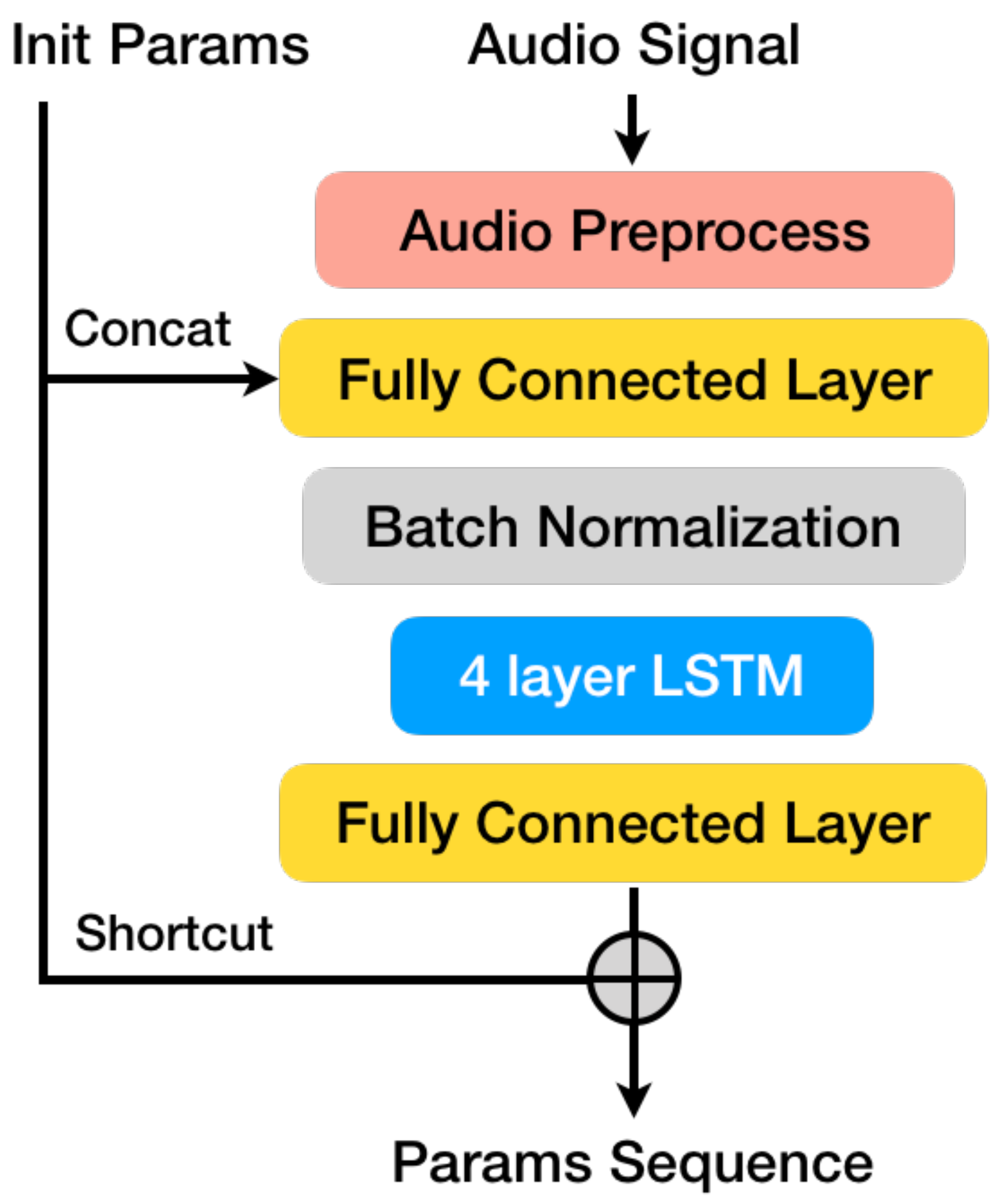}
	\caption{\textbf{Architecture of the sequential driver model.} For each frame, we approximate the residual relative to the initial parameters of the first frame~(reference image)}\label{fig:lstm}
\end{figure}

In the ViCo challenge~\cite{ViCo}, participants are restricted to training their methods on ViCo training set. We observe large performance differences between the models on the training and validation sets. To overcome this over-fitting issue, we introduce several model regularization techniques, including residual learning~\cite{he2016deep}, Dropout~\cite{srivastava2014dropout}, and training using BN~\cite{ioffe2015batch} with big batch size. The architecture of sequential driver containing a four-layer LSTM~\cite{hochreiter1997long} model is illustrated in Figure~\ref{fig:lstm}. 

\subsection{Learning}
\noindent\textbf{Loss Function}

For model optimization, we supervise each prediction using 3DMM parameters extracted from training videos. We randomly sample clips with $T=90$ frames and calculate the loss function as:

\begin{equation}
    \mathbf{L}_{gen} = \sum_{t=1}^{T} \Vert \beta_t - \widehat{\beta}_t \Vert_2 + \Vert c_t - \widehat{c}_t \Vert_2 + \Vert p_t - \widehat{p}_t \Vert_1,
\end{equation}

\noindent where $\beta, p, c$ denote the ground truth and $\widehat{\beta}, \widehat{p}, \widehat{c}$ represent the result of driver model. We experimentally search the choice of $L_1$ and $L_2$ loss functions. A head motion constraint loss $\mathbf{L}_{mot}$ is applied to encourage the inter-frame continuity:
\begin{equation}
    \mathbf{L}_{mot} = \sum_{t=1}^{T} \Vert \mu(c_t) - \mu(\widehat{c}_t) \Vert_2,
\end{equation}
\noindent where $\mu(\cdot)$ measures the inter-frame changes \ie $\mu(c_t)=c_t-c_{t-1}$. The final loss function can be formulated as:

\begin{equation}
    \mathbf{L}_{total} = \mathbf{L}_{gen} + \mathbf{L}_{mot}.
\end{equation}

\noindent\textbf{Details} 

For audio preprocessing, we follow the ViCo baseline~\cite{zhou2021responsive}. We extract the 14-dim Mel-frequency cepstral coefficients (MFCC) feature with the corresponding MFCC Delta and Delta-Delta~(second-order difference) feature. The energy, loudness and zero-crossing rate (ZCR) are also embedded into audio features $S_i$ for each audio frame clip $A_i$. The $45$-dimensional audio feature extracted from $A_s^t$ is denoted as $S_s^t = {s_1, ..., s_t}$. 

Our driver model is trained on four TITAN 2080Ti GPUs for about three hours. We use AdamW~\cite{kingma2014adam,loshchilov2018fixing} optimizer with a weight decay of $0.05$. Our training uses a batch size of $128$. We gradually reduce the learning rate from $5\times 10^{-3}$ to $10^{-4}$ using cosine annealing during the whole training process.

\subsection{Enhanced Renderer}

We use a controllable portrait image generation model, PIRenderer~\cite{ren2021pirenderer}, to convert 3DMM face parameters to video based on the reference image. PIRenderer employs a subset of 3DMM parameters as the head motion descriptor. Overall, PIRenderer works very well. However, two issues may degrade the image quality, including background distortion and image border artifacts.

\noindent\textbf{Foreground-Background Fusion} 

\begin{figure}[tb]
	\centering
	\includegraphics[width=8cm]{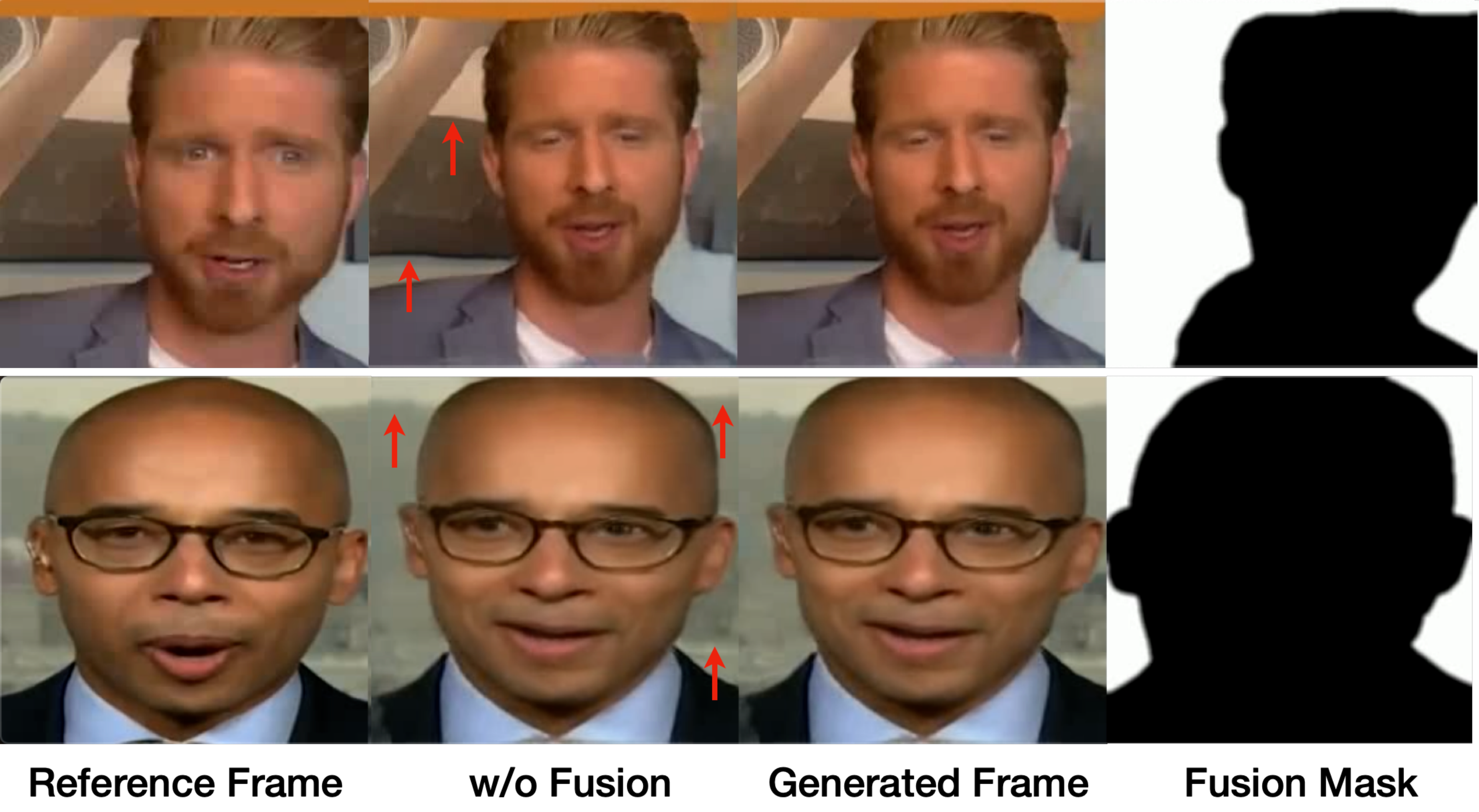}
	\caption{\textbf{The effect of foreground-background fusion.} The movement of the background is easily noticeable during video playback}\label{fig:fusion}
\end{figure}

Since we cannot estimate camera motion in the audio-driven head generation, we assume that the background is completely static. PIrenderer still needs background textures to complement some backgrounds revealed during head movement, which may not exist in the reference image. Due to the out-of-focus and complex texture of the background, the optical flow inferred by PIRenderer is not only on the surface of the head as wished. Therefore, we often observe unreasonable background distortions in synthetic videos, as shown in Figure~\ref{fig:fusion}.

To address this, We use a pre-trained foreground-background segmentation model, U$^2$Net~\cite{Qin_2020_PR}, to segment the static background area. For the generated image $\widehat{I}_t^{gen}$ and reference image $I^{ref}$, we detect their background regions:

\begin{equation}
M_t, M^{ref} = seg(\widehat{I}_t^{gen}), seg(I^{ref}),
\end{equation}

\noindent where $seg(\cdot)$ denotes the background region segmented from the image. To enhance inter-frame consistency, we calculate a median segmentation result, each pixel of which is the median of the results of the previous five frames. Formally, 

\begin{equation}
M^{med}_t(x, y) = median\{M_{t}(x, y), M_{t-1}(x, y), ..., M_{t-4}(x,y)\}, 
\end{equation}

\noindent where we calculate the recent median for each pixel-wise location.


Then we paste the common background area from reference image $I_{ref}$ to generated image $\widehat{I}_{gen}$:

\begin{equation}
M^{fusion}_t = Gaussian(M^{med}_t\cap M^{ref}),
\end{equation}
\begin{equation}
\widehat{I}_t^{fusion} = (1 - M^{fusion}_t) \odot \widehat{I}_t^{gen} + M^{fusion}_t \odot I^{ref},
\end{equation}

\noindent where we use a $7\times 7$ Gaussian filter to smooth the seam of the stitched image. 

\noindent\textbf{Image Boundary Inpainting} 

\begin{figure}[tb]
	\centering
	\includegraphics[width=8cm]{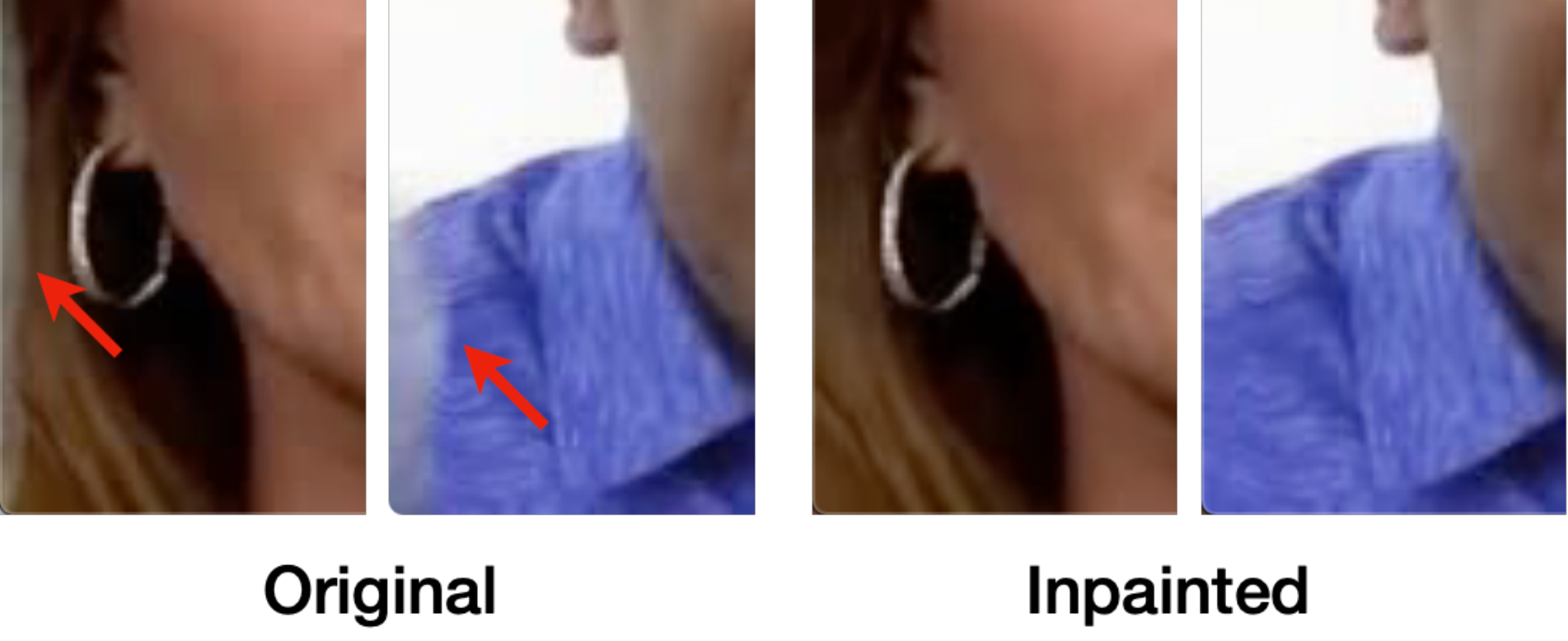}
	\caption{\textbf{The effect of image boundary inpainting.} When the features close to the boundary are warped, filling these holes with the boundary values of the original feature map will eliminate some artifacts}\label{fig:inpaint}
\end{figure}

For a good visual effect, the human hair and upper body should move with the movement of the head. In this case, we need to inpaint some texture around the edges of the image. We can easily perform this technique by setting the padding mode to ``border" in  \emph{grid\_sample} function of PyTorch~\cite{paszke2017automatic} when warping image features. The result is shown in Figure~\ref{fig:inpaint}.

Combining these two modules, we observe about $0.4$dB gain and $1.2$ drop of FID on our validation set without any extra training overhead.

\section{Experiments}

\subsection{Ablation Studies}
We do ablation studies on some designs, including regularization and model ensemble. The results are shown in Table~\ref{tab:ablation}.

\begin{table}[t]
	\caption{\textbf{The ablation studies on model design and training batch size}
	}
\label{tab:ablation}
\small
\resizebox{0.47\textwidth}{!}{\begin{tabular}{lccc}
\hline
\multicolumn{1}{c}{Setting}                         & \multicolumn{1}{c}{PSNR$\uparrow$}   & \multicolumn{1}{c}{FID$\downarrow$}     & \multicolumn{1}{c}{LMD$\downarrow$}            \\ \hline 
w/o BN~\cite{ioffe2015batch} & 16.16 & 17.67 &	50.00 \\
w/o residual learning~\cite{he2016deep} & 14.79	& 19.20 & 65.18 \\
w/o Dropout~\cite{srivastava2014dropout} & \textbf{16.38} & 17.12 & 44.34 \\
Final Model & 16.25 & \textbf{16.73} & \textbf{43.02} \\
\hline
batch size 8 & 16.52 & 18.13 & 43.07 \\
batch size 32 & \textbf{16.70} & 17.18 & \textbf{42.06} \\
batch size 128 & 16.25 & \textbf{16.73} & 43.02 \\
\hline 
\normalsize
\vspace{-1em}
\end{tabular}}
\end{table}

\noindent\textbf{Ablation on Model Design} 

Ablation experiments on model design are reported in Table~\ref{tab:ablation}.
Our model learns the residual added to the initial 3DMM parameters. Removing this design will greatly reduce the performance of the model~($1.46$dB). We also observe that BN and Dropout can help alleviate the over-fitting issue.

We further increase the batch size for training, specifically $128$. Our experiments show that large batch sizes have overall competitive performance while reducing training time exponentially.

\subsection{Model Ensemble}
We observe whether ensemble trained models in the same training process~(self ensemble) or models in different training processes~(cross-model ensemble), all performance metrics can be improved. The experiment results are shown in Table~\ref{tab:ensemble}. In the ViCo competition, we finally validate $10$ models on our validation set and cross-ensemble the best $5$ models. The most time spent in our system is in the PIRenderer part, so the ensemble on the driver model only slightly increases the inference overhead.

\begin{table}[t]
	\caption{\textbf{The ablation study on ensemble learning}
	}
\label{tab:ensemble}
\small
\resizebox{0.47\textwidth}{!}{\begin{tabular}{lccc}
\hline
\multicolumn{1}{c}{Setting}                         & \multicolumn{1}{c}{PSNR$\uparrow$}   & \multicolumn{1}{c}{FID$\downarrow$}     & \multicolumn{1}{c}{LMD$\downarrow$}             \\ \hline 
w/o ensemble & 16.25 & 16.73 & 43.02 \\
self ensemble~($3\times$) & 16.34 & 16.31 & 42.65 \\ 
cross-model ensemble~($3\times$) & \textbf{16.56} & \textbf{15.81} & \textbf{41.73} \\
\hline 
\normalsize
\vspace{-1em}
\end{tabular}
}
\end{table}

\section{Future Work}
We do not have enough time to fully explore the following relevant techniques in this short challenge, which might be very useful for this task. 1) Finetuning the renderer for current tasks and metrics is a reasonable point of improvement. The existing renderers produce results with inaccurate character identity retention and background disturbance. We could get a more targeted model to restrict the renderers to a specific application scenario, including rendering the same character and static background. 2) The lip shape generated by our talking head model is relatively conservative and insufficient to distinguish different syllables. The mapping of syllables to lips is a long-studied topic, and some experience from traditional methods may help to improve our model further. 3) There is also much room for exploration in more feature engineering techniques and stronger models.

\section{Conclusion}
In this paper, we introduce our solution for the conversational head generation competition. We propose a regularized driver and enhanced renderer to synthesize perceptually impressive videos. Hopefully, our discoveries and engineering practices can help future researchers.

\bibliographystyle{ACM-Reference-Format}
\balance
\bibliography{sample-base}










\end{document}